\RequirePackage{fix-cm}
\documentclass{article} 
\makeatletter
\def\input@path{{./style/}}
\makeatother

\usepackage[table]{xcolor}
\definecolor{loomrow}{HTML}{FFF8E5}
\definecolor{loopone}{HTML}{E7F1FB}
\usepackage{hyperref}

\usepackage[preprint]{report}

\usepackage{microtype}
\usepackage{roboto}
\usepackage{graphicx}
\usepackage{subcaption}
\graphicspath{{figures/}}
\usepackage{booktabs}
\usepackage{multirow}
\usepackage{amssymb}
\usepackage{pifont}
\usepackage{tcolorbox}
\tcbuselibrary{skins}

\usepackage{amsmath,amsfonts,bm}

\def\eqref#1{equation~\ref{#1}}

\def\1{\bm{1}}

\DeclareMathAlphabet{\mathsfit}{\encodingdefault}{\sfdefault}{m}{sl}
\SetMathAlphabet{\mathsfit}{bold}{\encodingdefault}{\sfdefault}{bx}{n}

\usepackage{algpseudocode}
\makeatletter
\algrenewcommand\ALG@beginalgorithmic{\setlength{\itemsep}{1pt}\setlength{\parsep}{0pt}}
\makeatother
\renewcommand{\algorithmiccomment}[1]{\hfill\(\triangleright\) #1}
\renewcommand{\Comment}[1]{\algorithmiccomment{#1}}
\algtext*{EndIf}
\algtext*{EndFor}

\newcommand{\method}{\texttt{LOOM}}
\newcounter{contribctr}
\newenvironment{contributions}{%
  \setcounter{contribctr}{0}%
  \begin{list}{}{%
    \setlength{\leftmargin}{1.7em}%
    \setlength{\labelwidth}{1.2em}%
    \setlength{\labelsep}{0.5em}%
    \setlength{\itemsep}{0.45em}%
    \setlength{\parsep}{0pt}%
    \setlength{\topsep}{0.35em}%
    \setlength{\partopsep}{0pt}%
  }%
}{\end{list}}
\newcommand{\citem}{%
  \stepcounter{contribctr}%
  \item[\ding{\numexpr181+\value{contribctr}\relax}]%
}

\icmltitlerunning{Looping Beyond Twice: A Scalable Recipe for Looped Mixture-of-Experts}

\definecolor{arxivcream}{RGB}{254,248,242}
\definecolor{arxivaffil}{RGB}{82,76,71}
\definecolor{arxivlink}{RGB}{16,16,130}
\makeatletter
\renewcommand{\printAffiliationsAndNotice}[1]{\global\icml@noticeprintedtrue}
\makeatother

\begin{document}

\thispagestyle{plain}
\vspace*{-1cm}
\begin{tcolorbox}[
  enhanced,
  colback=arxivcream,
  colframe=arxivcream,
  boxrule=0pt,
  arc=7pt,
  left=16pt, right=16pt, top=14pt, bottom=12pt,
  before skip=0pt, after skip=12pt,
  width=\linewidth
]
\begin{center}
\setlength{\parskip}{0pt}%
{\robotoblack\fontsize{17.5}{19.4}\selectfont
\raisebox{-0.4em}{\includegraphics[height=1.5em]{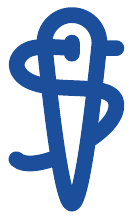}}\enspace
Looping Beyond Twice: A Scalable Recipe for Looped Mixture-of-Experts\par}
\vspace{14pt}
{\robotobold\fontsize{11}{12.6}\selectfont
Di He\textsuperscript{1,2,3},
Pengxiang Li\textsuperscript{4},
Da Chang\textsuperscript{1,2,3},
Qingyan Meng\textsuperscript{2},
Lu Yin\textsuperscript{5},
Shiwei Liu\textsuperscript{6,7,8,\dag}\par}
\vspace{12pt}
{\robotoregular\fontsize{8.6}{12.8}\selectfont\color{arxivaffil}
\textsuperscript{1}Shenzhen Institutes of Advanced Technology, Chinese Academy of Sciences,\par
\textsuperscript{2}Peng Cheng Laboratory,
\textsuperscript{3}University of Chinese Academy of Sciences,\par
\textsuperscript{4}The Hong Kong Polytechnic University,
\textsuperscript{5}University of Surrey,
\textsuperscript{6}ELLIS Institute T\"ubingen,\par
\textsuperscript{7}Max Planck Institute for Intelligent Systems,
\textsuperscript{8}T\"ubingen AI Center\par
\textsuperscript{\dag}Corresponding author: sliu@tue.ellis.eu\par}
\end{center}
\vspace{10pt}
{\rmfamily\fontsize{10.2}{14.6}\selectfont
\setlength{\parindent}{0pt}%
Looped Transformers introduce recurrent depth as a new scaling axis for large language models (LLMs): by repeatedly applying shared Transformer blocks, they increase effective depth without increasing parameter count. However, the benefits of looping remain unclear for large MoE LLMs under FLOPs-matched comparisons. The main reason is that the gains from additional iterations diminish quickly and can even turn into degradation, so the extra FLOPs spent on looping yield little substantial improvement. Consequently, prior work typically settles on two loops. In this paper, we take a closer look at why looped MoE fails to scale and identify two main obstacles. First, looping inherits and amplifies the \emph{curse of depth}: hidden-state variance grows with each iteration as residual updates accumulate, which destabilizes deep recurrence and causes representations to drift. Second, looped MoE suffers from \emph{expert selection collapse}: routers repeatedly select the same experts across loops, so extra iterations add computation without adding computational diversity. Guided by this diagnosis, we propose {\method}, built on a single principle: \emph{each loop should contribute new computation while keeping the recurrent state stable.} {\method} \emph{stabilizes} recurrence by scaling residual updates to bound
variance growth and re-injecting the input embedding at every loop, and
\emph{diversifies} it through per-loop routers that engage different experts
and a Looping Residual that carries earlier outputs forward.  Experiments across 100M--1.7B models show stable scaling to 9--12 loops, with the 1.7B model scaling to 9 loops. \textbf{Under near-iso-FLOP, the 700M model performs best at 5 loops},
reducing perplexity from $18.36$ to $16.54$ and improving average zero-shot
accuracy from $38.84\%$ to $39.53\%$ over the non-looped baseline.
\textbf{Without FLOP matching, the 1.7B model trained on 60B tokens peaks at
9 loops}, reducing perplexity from $9.62$ to $7.77$ and improving
average zero-shot accuracy from $42.4\%$ to $47.7\%$.
\textbf{Code:} {\color{arxivlink}\href{https://github.com/hed-ucas/LOOM}{LOOM}}\par}
\end{tcolorbox}
\printAffiliationsAndNotice{}

\begingroup
\setlength{\intextsep}{0pt}%
\setlength{\abovecaptionskip}{3pt}%
\setlength{\belowcaptionskip}{0pt}%
\newlength{\loomrest}%
\newlength{\loompanel}%
\setlength{\loomrest}{\dimexpr\textheight-\pagetotal-8pt\relax}%
\newcommand{\loomscale}{0.875}
\setlength{\loompanel}{\dimexpr(\linewidth-12pt)*1000/2857\relax}%
\setlength{\loompanel}{\loomscale\loompanel}%
\ifdim\loompanel>\dimexpr\loomrest-40pt\relax
  \setlength{\loompanel}{\dimexpr\loomrest-40pt\relax}%
\fi
\typeout{LOOM page1 rest=\the\loomrest\space panel=\the\loompanel\space pagetotal=\the\pagetotal}%
\vspace*{\fill}%

\begin{figure}[H]
\centering
\captionsetup{skip=2pt,font=small}
\includegraphics[height=\loompanel]{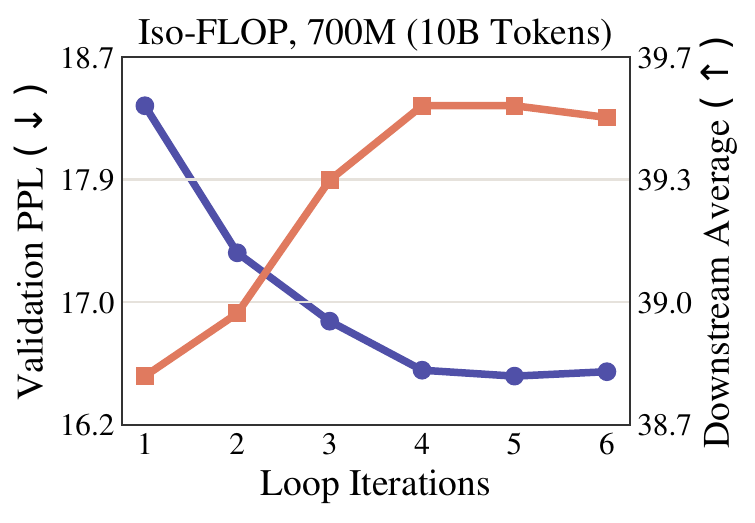}\hspace{12pt}%
\includegraphics[height=\loompanel]{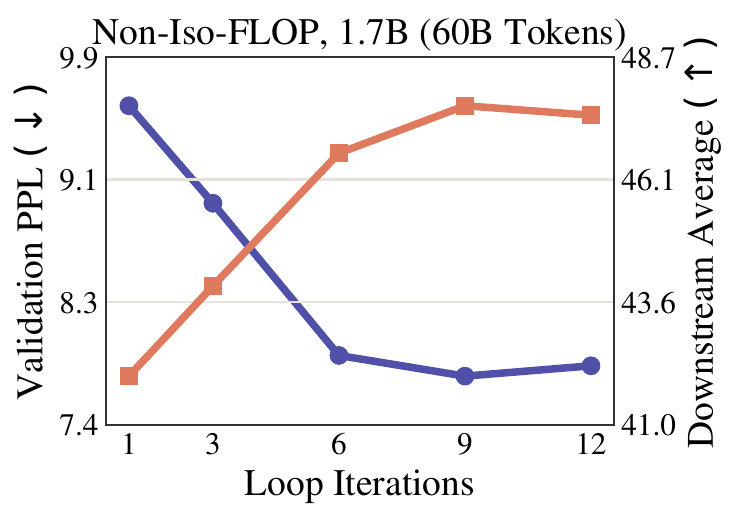}
\includegraphics[width=0.55\linewidth]{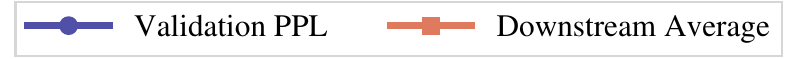}
\caption{\textbf{Loop-depth scaling of {\method}.}
\textbf{Left:} under a near-iso-FLOP budget, {\method} achieves its best performance near 5 loops.
\textbf{Right:} without FLOP matching, the 1.7B model scales to 9 loops, with substantial performance improvements.}
\label{fig:loopsweep}
\end{figure}

\endgroup
\clearpage

\vspace{-0.1cm}
\section{Introduction}
\label{sec:intro}
\vspace{-0.1cm}

Looped Transformers scale the computational depth of LLMs by repeatedly applying shared Transformer blocks, increasing effective depth without increasing parameters~\citep{dehghani2018universal,giannou2023looped,saunshi2025reasoning,geiping2026scaling}. Unlike conventional depth scaling, looping introduces an additional scaling axis orthogonal to model size and training data~\citep{kaplan2020scaling,hoffmann2022training}. Recent studies demonstrate that recurrent depth improves algorithmic computation~\citep{giannou2023looped}, length generalization~\citep{fan2025looped}, and latent reasoning~\citep{saunshi2025reasoning,geiping2026scaling,hao2024training}, establishing looped computation as a promising direction for LLM scaling.


MoE architectures offer a natural pathway for extending looped computation. By activating only a sparse subset of experts per token, MoE models provide additional computation capacity for recurrent refinement under a fixed budget~\citep{shazeer2017outrageously,fedus2022switch}. However, current looped MoE systems struggle to benefit from deeper recurrence: increasing the number of loops often leads to loss degradation rather than further improvements. Beyond this regime, additional iterations yield diminishing returns. This has raised concerns about the practical benefits of looped MoE under matched FLOPs, since further increasing the number of loops mainly adds computation without a corresponding performance gain. Consequently, existing approaches often limit recurrence to two iterations~\citep{gao2026loop,wang2026smelt,lab2026nanbeige4}.

\vspace{-0.4cm}
\begin{figure}[H]
\centering
\includegraphics[width=0.8\linewidth]{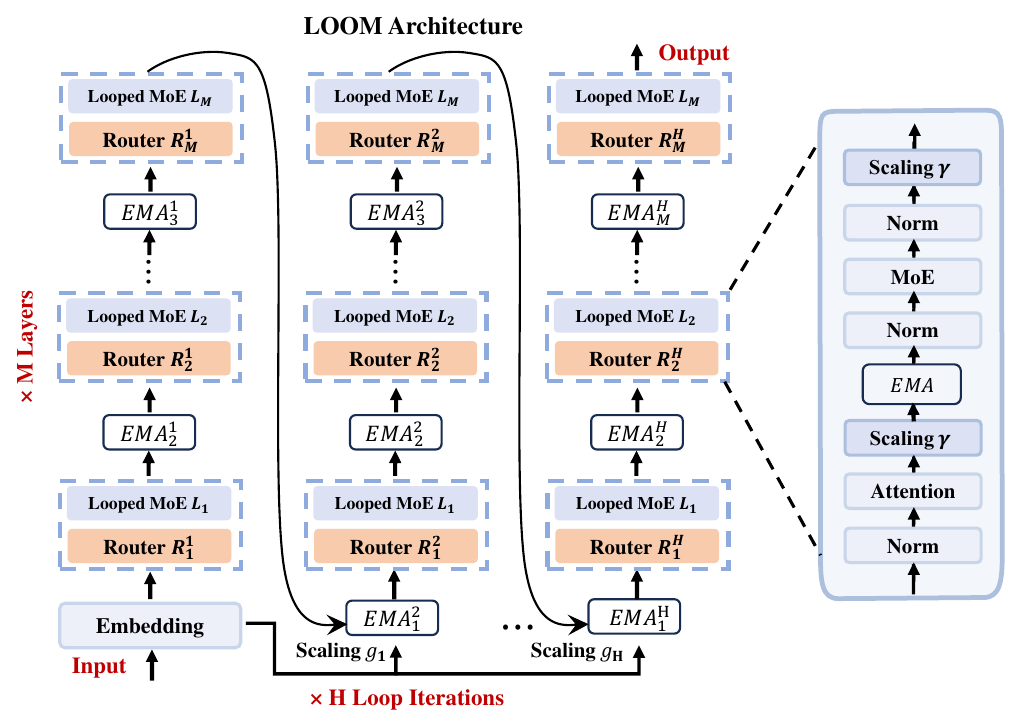}
\caption{{\method} architecture. An $M$-layer MoE block is reused for $H$ loops, with $\lambda$ as a scalar hyperparameter. At layer $a$ of loop $t$, $\gamma=\lambda/(H\sqrt{M})$ scales residual updates, $g_t=\lambda/(t\sqrt{M})$ controls embedding re-injection, $\mathrm{EMA}_a^t$ denotes the Looping Residual, and $R_a^t$ denotes the router.}
\label{fig:method}
\vspace{-0.5cm}
\end{figure}

To understand why deeper recurrence fails to pay off, we investigate the root causes of this limitation and identify two main obstacles to scaling looped MoE. First, looping amplifies the \emph{curse of depth \cite{sun2026curse}}: repeatedly reusing shared Transformer blocks causes residual accumulation and activation variance growth, as shown in Figure~\ref{fig:resembed}, which makes hidden states drift and destabilizes optimization in deep recurrence. Second, looped MoE suffers from \emph{expert selection collapse}: shared routers produce nearly invariant expert assignments across loops, so later iterations repeatedly activate similar experts instead of refining earlier representations (shown in Figure~\ref{fig:routing}). As a result, increasing loop depth adds computation without adding useful computation. This reveals the central challenge of deep looped MoE: 
\begin{center}
\begin{tcolorbox}[width=0.9\textwidth]
\textit{Is there an effective way to improve cross-loop information flow in looped MoE, so that recurrence can scale well beyond two iterations?}
\end{tcolorbox}
\end{center}

To answer this question, we propose {\method}, built on a single principle: \emph{each loop should contribute new computation while keeping the recurrent state stable.} {\method} realizes this principle along two complementary axes, each targeting one of the obstacles identified above.

\textbf{Diversifying recurrence.} To counter expert selection collapse, {\method} equips each loop with its own router, so that different iterations engage different subsets of experts rather than repeatedly activating the same ones. To make these diverse computations build on one another, {\method} further introduces a \emph{Looping Residual}, which carries the outputs of earlier loops forward so that later iterations can refine previous representations rather than overwrite them.

\textbf{Stabilizing recurrence.} To counter the curse of depth, {\method} adopts residual scaling~\citep{sun2026curse,noci2022signal,wang2026residual}, scaling the output of each attention and MoE block by $\gamma = \lambda / (H\sqrt{M})$. This bounds residual accumulation and keeps signal propagation stable as the number of loops grows. In addition, to prevent deep recurrence from diluting input information and causing latent drift, {\method} re-injects the input embedding at each iteration~\citep{geiping2026scaling}. To adapt this technique to looped MoE, we recalibrate the injected embedding at step $t$ with $g_t = \lambda / (t\sqrt{M})$, so that the anchoring signal remains well-scaled across iterations.

Together, these components enable shared experts to progressively refine representations across loops, allowing {\method} to scale recurrence well beyond two iterations without routing collapse or recurrent degradation. Our main contributions are as follows:

\begin{contributions}
\citem We identify two main obstacles that prevent looped MoE from scaling beyond two iterations: the \emph{curse of depth}, in which residual accumulation causes variance growth and latent drift, and \emph{expert selection collapse}, in which shared routers assign nearly identical experts across loops.
\citem We propose {\method}, a principled recipe for deep looped MoE built on a single design principle: each loop should contribute new computation while keeping the recurrent state stable. {\method} stabilizes recurrence through residual scaling and embedding re-injection, and diversifies recurrence through loop-specific routers and a Looping Residual.
\citem Experiments across model scales from 100M to 1.7B parameters and up to 60B training tokens show that {\method} scales to 9--12 loops, and that the 1.7B model scales to 9 loops. Notably, under matched FLOPs, {\method} scales looping to 5 iterations while consistently achieving lower loss, providing strong evidence that recurrent depth is a genuine and compute-efficient scaling axis for MoE LLMs.
\end{contributions}

\vspace{-0.4cm}
\section{Related Work}
\label{sec:related}
\vspace{-0.1cm}

\textbf{Looped models.}
Looped LLMs reuse the same Transformer blocks along the depth axis, increasing effective depth without increasing parameter count. This idea traces back to Universal Transformer~\citep{dehghani2018universal} and ALBERT~\citep{lan2019albert}. Recent studies demonstrate the potential of recurrent computation in dense Transformers: \citet{geiping2026scaling} show that test-time looping improves language models; \citet{saunshi2025reasoning} connect recurrence with latent-space reasoning; and \citet{fu2026simply}, \citet{wang2026residual}, and \citet{movahedi2026fixed} enable deeper recurrence through stabilization, residual scaling, and fixed-point formulations. HRM and TRM further explore recursive architectures for iterative reasoning~\citep{wang2025hierarchicalreasoningmodel,jolicoeur2025less}. Chain-of-Experts studies iterative expert computation within each layer rather than recurrently reusing an entire shared MoE block~\citep{wang2025chain}. Extending recurrent computation to sparse MoEs introduces additional challenges due to routing dynamics and cross-loop information propagation.

\textbf{Stability in deep residual stacks.}
Deep recurrence increases effective depth but can amplify residual dynamics, causing activation instability and loss degradation. \citet{noci2022signal} show that rank collapse of token representations causes vanishing gradients in Transformers and can be prevented by scaling residual branches, while \citet{sun2026curse} identify the curse of depth in LLMs, where Pre-LN variance growth renders deeper layers nearly identity mappings and propose $1/(\sqrt{l})$ to scale down the residual. To address this, Mix-LN~\citep{li2024mix} applies Post-LN to earlier layers and Pre-LN to deeper layers. Another line of work redesigns the residual pathway itself: mHC~\citep{xie2025mhc} constrains the residual mixing matrices of hyper-connections to a doubly stochastic manifold to restore the identity-mapping property and stabilize large-scale training, and Attention Residuals~\citep{team2026attention} replace fixed additive accumulation with attention over earlier layer outputs, allowing each layer to selectively aggregate previous representations. Residual scaling has also been applied to recurrent models: \citet{wang2026residual} introduce a loop-aware factor $\lambda/(H\sqrt{M})$, and \citet{movahedi2026fixed} combine residual scaling with inter-loop mixing. Input re-injection further alleviates representation drift~\citep{geiping2026scaling}. These techniques lay the foundation for stabilizing deep recurrence and motivate their adaptation to looped MoE.

\textbf{Looped MoE.}
Sparse gating enables MoE models to scale parameter capacity under limited computation, but introduces additional challenges for recurrent computation, particularly in expert routing and cross-loop information propagation~\citep{shazeer2017outrageously,fedus2022switch}. Existing recurrent MoE studies have explored limited forms of recurrence. MoEUT extends Universal Transformer with MoE layers rather than stabilizing deep full-block recurrence~\citep{csordas2024moeut}; Chain-of-Experts iterates experts within layers instead of reusing a shared MoE block~\citep{wang2025chain}; SMELT studies compute-matched MoE looping and identifies two loops as best~\citep{wang2026smelt}. LoopMoE introduces iteration-adaptive normalization, while \citet{lee2026sparse} study recurrent MoE models with varying expert utilization but maintain effective depth by reducing layers as loop counts increase~\citep{chen2026loopmoe,lee2026sparse}. These approaches demonstrate the potential of recurrent MoE, but scalable deep recurrence under a fixed architecture remains underexplored.\vspace{-0.1cm}

\vspace{-0.2cm}
\section{Methodology}
\label{sec:method}
\vspace{-0.1cm}

\paragraph{Overview.}
LOOM repeatedly applies an $M$-layer, pre-norm MoE decoder. Each loop applies layers $\ell=1,2,\dots,M$ in order and repeats this walk for $H$ passes. The attention and expert weights are shared across loops, whereas each loop has its own routers. The resulting effective depth is $MH$; increasing $H$ adds router parameters but does not replicate the attention or experts. Let $x=\operatorname{Embed}(\mathrm{tokens})$ and let $h_\ell^t$ denote the residual stream after layer $\ell$ of loop $t$, with $h_0^1=x$. The final state $h_M^H$ is used for next-token prediction.

\vspace{-0.1cm}
\subsection{Stabilizing the recurrent state}\label{sec:method_stability}
\vspace{-0.1cm}

\begin{figure}[t]
\centering
\captionsetup{skip=4pt,font=small}
\captionsetup[subfigure]{skip=1pt,font=small}
\begin{minipage}{\linewidth}
\centering
\includegraphics[width=0.74\linewidth]{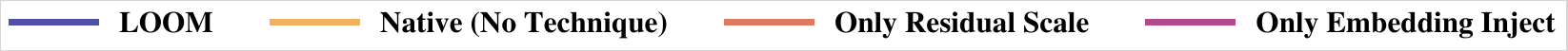}
\par\vspace{1.5mm}
\begin{subfigure}[t]{0.318\linewidth}
\centering
\includegraphics[width=\linewidth]{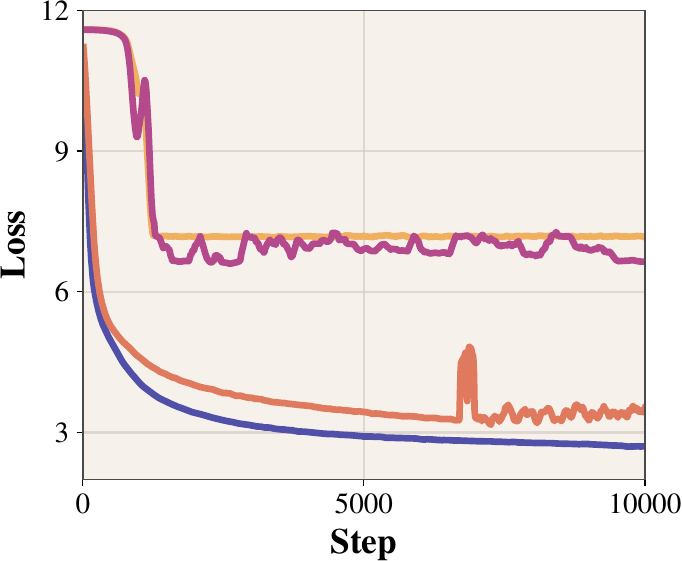}
\end{subfigure}\hspace{2.2mm}%
\begin{subfigure}[t]{0.318\linewidth}
\centering
\includegraphics[width=\linewidth]{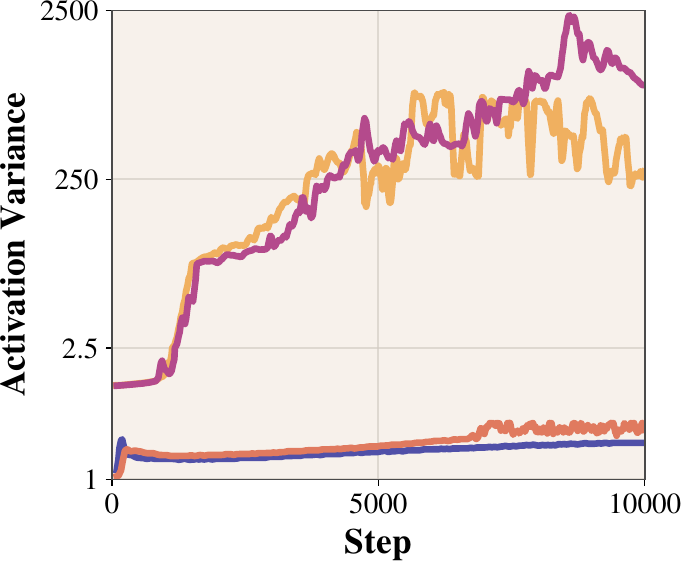}
\end{subfigure}\hspace{2.2mm}%
\begin{subfigure}[t]{0.318\linewidth}
\centering
\includegraphics[width=\linewidth]{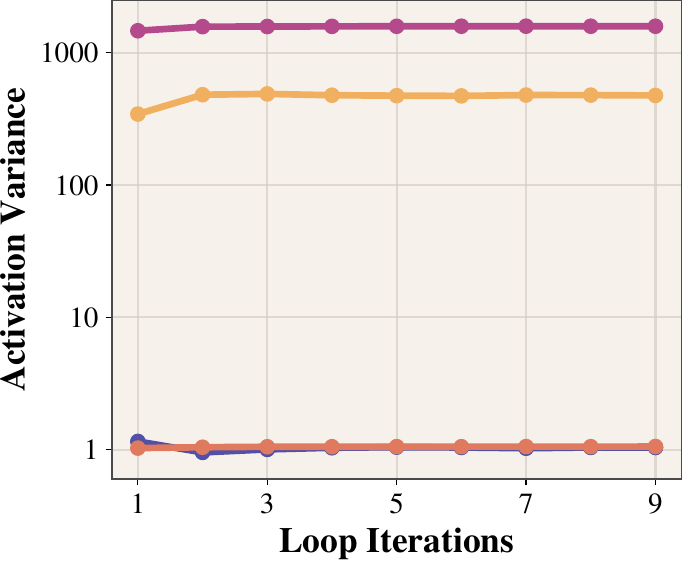}
\end{subfigure}
\end{minipage}
\caption{Effect of residual scaling and embedding re-injection on training stability in a 9-loop MoE model (${\sim}350$M parameters, $10$B training tokens). \textbf{Native (No Tech)} loops the unmodified backbone without additional techniques. \textbf{Only Res Scale} and \textbf{Only Embed Inject} add residual scaling and embedding re-injection, respectively, to Native, with no other modifications. (a)~Training loss. (b)~Activation variance during training. (c)~Activation variance across recurrent iterations.}
\label{fig:resembed}
\vspace{-0.5cm}
\end{figure}

\paragraph{Residual scaling.}
Deep recurrence amplifies the curse of depth~\citep{sun2026curse}: correlated residual updates accumulate across loops, increasing hidden-state variance and destabilizing signal propagation. Figure~\ref{fig:resembed}(b,c) shows that unscaled residual accumulation leads to activation variance explosion during training and across loops, whereas residual scaling effectively controls this growth. Accordingly, {\method} incorporates residual scaling that accounts for both loop count and physical depth~\citep{wang2026residual}:
\begin{equation}
\gamma=\frac{\lambda}{H\sqrt{M}},\quad
\widetilde h_\ell^t=h_{\ell-1}^t+\gamma\,\mathrm{Attn}_\ell(\mathrm{RMS}(h_{\ell-1}^t)),\quad
h_\ell^t=\widetilde h_\ell^t+\gamma\,\mathrm{RMS}(\mathrm{MoE}_\ell^t(\mathrm{RMS}(\widetilde h_\ell^t))).
\label{eq:gamma}
\end{equation}

The $1/H$ factor limits residual accumulation across loops, while $1/\sqrt M$ controls variance growth across physical layers. {\method} fixes $\lambda=0.5$ as a non-learnable hyperparameter and applies $\gamma$ uniformly across layers and loops. Scaling the branch outputs preserves the identity path, while the additional MoE output normalization controls the magnitude of its residual contribution.

\paragraph{Embedding re-injection.} \label{sec:inject}
Residual scaling alone substantially reduces activation variance but yields higher training loss than the complete recipe (Figure~\ref{fig:resembed}), indicating that variance control alone is insufficient. Deep recurrence can also dilute input information and induce representation drift~\citep{wang2026residual}. To preserve access to the original token representation, {\method} re-injects the input embedding at the beginning of each loop $t\geq2$:
\begin{equation}
    g_t=\frac{\lambda}{t\sqrt{M}}.
        \qquad
    h_0^t=(1-g_t)h_M^{t-1}+g_t x,
    \label{eq:embedding_reinjection}
\end{equation}
We use the same fixed $\lambda=0.5$ as in residual scaling. Re-injection begins at $t=2$, since $h_0^1=x$. The decreasing coefficient $g_t$ places progressively greater weight on the recurrent representation, while convex mixing regulates the embedding contribution. Removing embedding re-injection from {\method} increases validation perplexity from $18.62$ to $26.05$ at $H=9$ (Table~\ref{tab:ablate}), supporting its complementary role in anchoring recurrent states alongside residual scaling.

\begin{figure}[t]
\centering
\captionsetup{skip=4pt,font=small}
\captionsetup[subfigure]{skip=1pt,font=small}
\resizebox{\linewidth}{!}{%
\includegraphics[height=4.2cm]{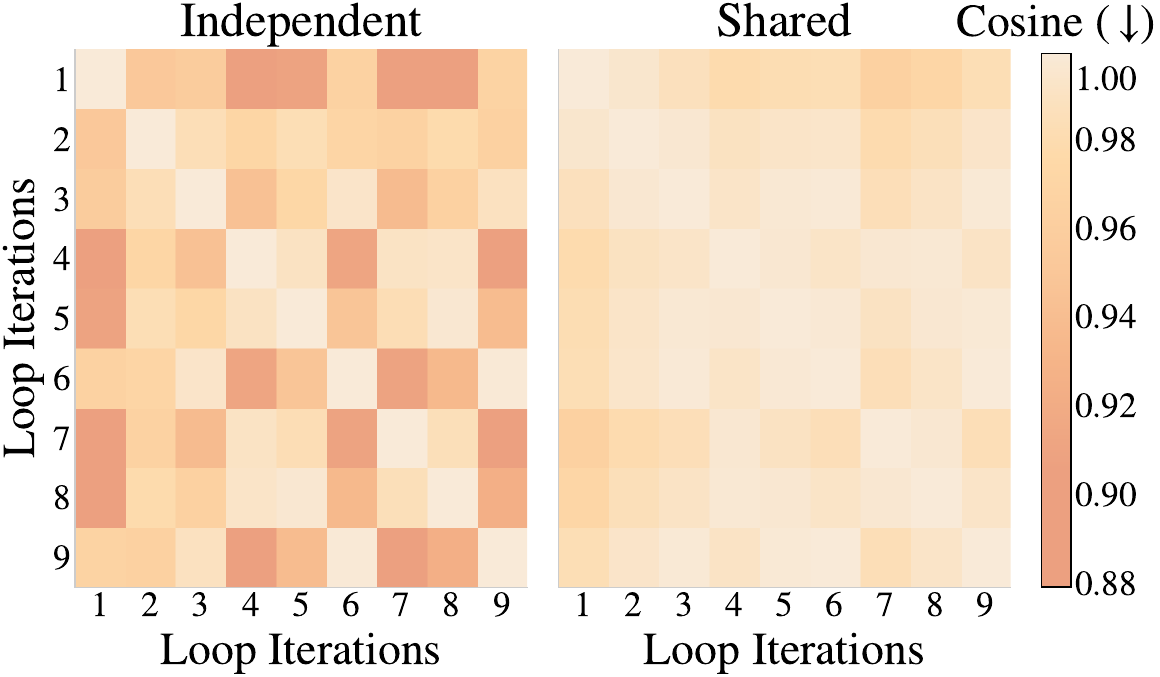}%
\hspace{2.5mm}%
\includegraphics[height=4.2cm]{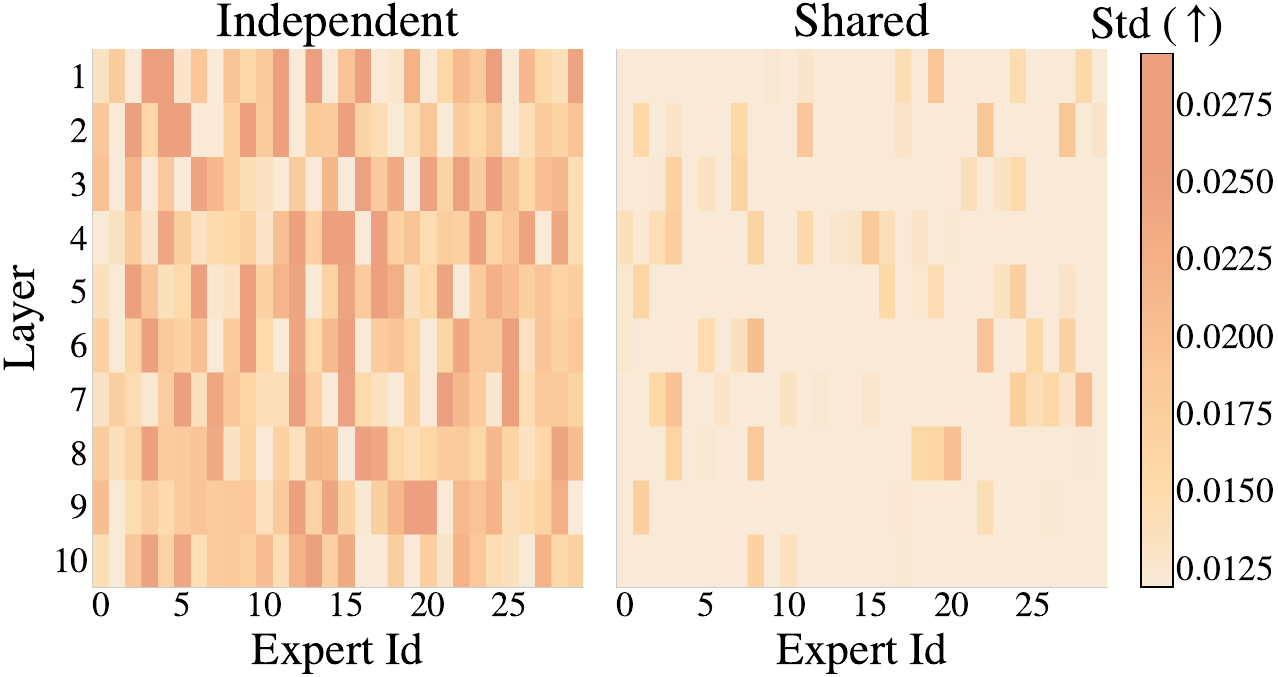}%
}
\begin{subfigure}[t]{0.49\linewidth}
\label{fig:loopcos}
\end{subfigure}\hfill
\begin{subfigure}[t]{0.49\linewidth}
\label{fig:expertstd}
\end{subfigure}
\vspace{-2.5mm}%
\caption{Expert routing in a 9-loop MoE model (${\sim}350$M parameters), comparing loop-specific (Independent) and shared (Shared) routers. (a) Cross-loop similarity: $\cos(\bar{\mathbf{p}}_t,\bar{\mathbf{p}}_{t\prime})$, where $\bar{\mathbf{p}}_t$ is the normalized expert-load distribution averaged across layers for loop $t$. Each cell compares two loops; higher values indicate more similar expert utilization. (b) Cross-loop variability: $\operatorname{std}_t(p_{\ell,t,e})$, where $p_{\ell,t,e}$ is the normalized load of expert $e$ at layer $\ell$ in loop $t$. Each cell measures an expert's load variation across loops at a given layer. Loop-specific routers yield lower similarity and greater variability, indicating more diverse expert utilization.}
\label{fig:routing}
\vspace{-0.55cm}
\end{figure}

\vspace{-0.1cm}
\subsection{Diversifying computation across loops}
\vspace{-0.1cm}

\paragraph{Loop-specific routers.} \label{sec:router}
Deep recurrence should enable successive loops to perform complementary computations. However, sharing routers across loops can produce nearly invariant expert selections, leading to \emph{expert selection collapse} and limiting the benefits of additional iterations. {\method} addresses this limitation through \textbf{loop-specific routing with shared experts}: each layer uses a separate router for each loop, while its expert weights remain shared. This decouples routing decisions across iterations, enabling different stages of recurrence to access different expert combinations without replicating expert parameters. Figure~\ref{fig:routing} provides two complementary observations: \textbf{(i)} lower cross-loop cosine similarity of expert-load distributions and \textbf{(ii)} greater variation in expert loads across loops. Together, these observations support more diverse expert utilization, a prerequisite for complementary computation across recurrent iterations.

\paragraph{Looping Residual.} \label{sec:attnres}
Loop-specific routers diversify computation across iterations, while progressive refinement additionally requires effective reuse of earlier computations. {\method} therefore introduces a \textbf{Looping Residual} based on a \textbf{fixed-decay exponential moving average (EMA)}, with two complementary memories: \textbf{(i)} a global memory spanning layers and loops, and \textbf{(ii)} a local memory aggregating attention outputs within the current loop. For layer $\ell$ in loop $t$, let
$o_\ell^t=\gamma\,\mathrm{Attn}_\ell(\mathrm{RMS}(h_{\ell-1}^t))$
denote the scaled attention output. Each memory maintains an accumulator $N_A$ and a normalization factor $D_A$:
\begin{equation}
N_A\leftarrow\beta N_A+o_\ell^t,\quad
D_A\leftarrow\beta D_A+1,\quad
r_A=\frac{N_A}{D_A},
\qquad A\in\{H,L\}.
\label{eq:lr_memory}
\end{equation}
Both pairs are initialized to zero, with fixed $\beta=0.5$. The global pair persists across the full recurrence, whereas the local pair is reset at the beginning of each loop. Normalization by $D_A$ yields exponentially weighted summaries that emphasize recent computations while retaining earlier contributions. Crucially, \textbf{each memory stores only one accumulator tensor and one scalar, regardless of the number of loops}. This recursive formulation avoids storing the output history and requires only elementwise operations per update, introducing negligible additional memory and computational overhead relative to the backbone. The summaries replace the direct attention residual update in Equation \ref{eq:gamma}:
\begin{equation}
\widetilde h_\ell^t=h_{\ell-1}^t+r_H+r_L.
\label{eq:looping_residual}
\end{equation}
The global summary $r_H$ and local summary $r_L$ aggregate attention outputs across loops and within the current loop, respectively, preserving intermediate computations for recurrent refinement. The resulting state $\widetilde h_\ell^t$ is subsequently processed by the MoE residual update defined in Eq.~\eqref{eq:gamma}.

\begin{figure}[t]
\centering
\captionsetup{skip=3pt,font=small}
\includegraphics[height=0.292\linewidth]{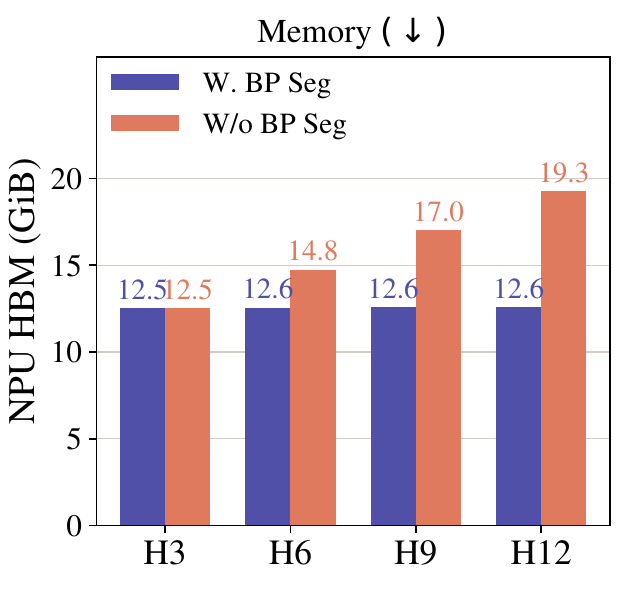}\hspace{0.6mm}%
\includegraphics[height=0.292\linewidth]{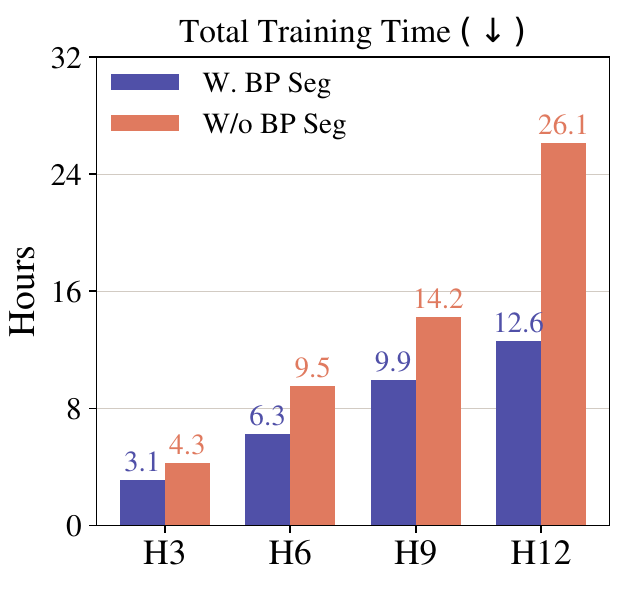}\hspace{0.6mm}%
\includegraphics[height=0.292\linewidth]{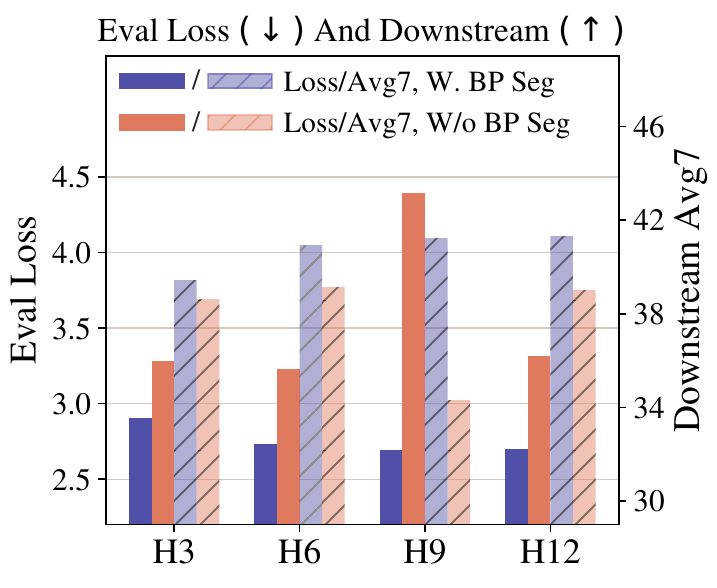}
\caption{Effect of segmented backpropagation ($K{=}3$) on the ${\sim}350$M model, compared with backpropagation through all loops. (a) Peak NPU memory measured with a microbatch size of 1. (b) Total training times. (c) Held-out evaluation loss (solid bars) and average accuracy across seven downstream tasks (hatched bars).}
\label{fig:bpseg}
\vspace{-0.3cm}
\end{figure}

\vspace{-0.1cm}
\subsection{Training with segmented backpropagation}
\label{sec:bpseg}
\vspace{-0.1cm}

{\method} uses segmented backpropagation, dividing the $H$ recurrent loops into consecutive segments of at most $K=3$ loops. Each segment receives supervision from a language-modeling loss at its final loop, with gradients restricted to that segment. The residual state and global Looping Residual memory are then propagated to the next segment with gradients detached, preserving forward information across all $H$ loops while limiting backpropagation depth.

Figure~\ref{fig:bpseg} compares this strategy with backpropagation through all $H$ loops on the ${\sim}350$M model, highlighting two benefits. \ding{182} Lower memory consumption and faster training. From $H=3$ to $H=12$, segmentation maintains peak memory near $12.6$\,GiB, whereas unsegmented memory increases from $12.5$ to $19.3$\,GiB. At $H=12$, the reported training time decreases from $26.1$ to $12.6$ hours. \ding{183} Improved stability and performance. Segmented runs achieve lower evaluation loss and higher average accuracy across seven downstream tasks. At $H=9$, the unsegmented run exhibits severe degradation, with evaluation loss reaching approximately $4.4$ and average accuracy falling to approximately $34\%$. These benefits in speed and stability motivate using segmented backpropagation with $K{=}3$ by default in {\method}. Detailed comparisons in Table~\ref{tab:bpk} further support using segmented backpropagation by default in {\method}.

\begin{algorithm}[t]
\caption{Training procedure of {\method}.
$\mathrm{sg}(\cdot)$ stops gradient flow.}
\label{alg:loom}
\begin{algorithmic}[1]
\Require Loop count $H$, depth $M$, segment length $K=3$, $\lambda=\beta=0.5$
\State $x,h_0^1\gets\mathrm{Embed}(\mathrm{tokens})$;
       $\gamma\gets\lambda/(H\sqrt{M})$;
        init shared Looping Residual $(N_H,D_H)\gets(0,0)$
\For{$t=1,\dots,H$}
  \State \textbf{if} $t\geq2$ \textbf{then}
         $g_t\gets\lambda/(t\sqrt{M})$;
         $h_0^t\gets(1-g_t)h_M^{t-1}+g_t x$
         \Comment{Formula (\ref{eq:embedding_reinjection})}
  \State $(N_L,D_L)\gets(0,0)$
         \Comment{Reset local memory}
  \For{$\ell=1,\dots,M$}
    \State $o_\ell^t\gets\gamma\,\mathrm{Attn}_\ell(\mathrm{RMS}(h_{\ell-1}^t))$;
           update both Looping Residual via~Formula (\ref{eq:lr_memory})
    \State $\widetilde h_\ell^t\gets h_{\ell-1}^t+r_H+r_L$;
           $h_\ell^t\gets\widetilde h_\ell^t+
           \gamma\,\mathrm{RMS}(\mathrm{MoE}_\ell^t(\mathrm{RMS}(\widetilde h_\ell^t)))$
  \EndFor
  \If{$t\bmod K=0$ \textbf{or} $t=H$}
    \State $\hat y^t\gets\mathrm{LMHead}(\mathrm{RMS}(h_M^t))$;
           backward; optimizer step
    \State $(h_M^t,N_H,D_H)\gets\mathrm{sg}(h_M^t,N_H,D_H)$
           \Comment{Stop gradient}
  \EndIf
\EndFor
\end{algorithmic}
\end{algorithm}

Algorithm~\ref{alg:loom} summarizes the forward computation and segmented training procedure of {\method}.

\vspace{-0.2cm}
\section{Empirical results}
\label{sec:exp}
\vspace{-0.2cm}

In this section, we present the experimental setup (Section~\ref{sec:setup}), loop-depth results from ${\sim}100$M to ${\sim}1.7$B (Section~\ref{sec:depth}), and ablations of the recipe (Section~\ref{sec:ablate350}).

\subsection{Experimental setup}
\label{sec:setup}

\vspace{-0.1cm}
\begin{table}[!t]
\centering
\caption{Architecture and optimization settings for the three {\method} scales. Effective depth is $MH$, with an $M$-layer block reused for $H$ recurrent iterations.}
\label{tab:train}
\label{tab:arch}
\small
\setlength{\tabcolsep}{3.5pt}
\resizebox{0.7\linewidth}{!}{%
\begin{tabular}{cccc}
\toprule
Parameters (activated) & ${\sim}100$M (${\sim}70$M) & ${\sim}350$M (${\sim}141$M) & ${\sim}1.7$B (${\sim}0.63$B) \\
\midrule
Physical layers $M$ & $6$ & $10$ & $15$ \\
Effective depth & $6H$ & $10H$ & $15H$ \\
Hidden size $d$ & $384$ & $512$ & $1280$ \\
Query / KV heads & $6$ / $3$ & $8$ / $4$ & $20$ / $10$ \\
Head dimension & $64$ & $64$ & $64$ \\
Expert width $I$ & $384$ & $512$ & $768$ \\
Routed / shared experts & $24$ / $2$ & $30$ / $2$ & $30$ / $2$ \\
Top-$k$ & $6$ ($4{+}2$) & $8$ ($6{+}2$) & $8$ ($6{+}2$) \\
Router & per-loop sigmoid, float32 & per-loop sigmoid, float32 & per-loop sigmoid, float32 \\
$z$-loss / capacity & $10^{-3}$ / $1.5$ & $10^{-3}$ / $1.5$ & $10^{-3}$ / $1.5$ \\
Load-balance bias LR & $2{\times}10^{-3}$ & $2{\times}10^{-3}$ & $2{\times}10^{-3}$ \\
$\lambda/\beta/K$ & $0.5/0.5/3$ & $0.5/0.5/3$ & $0.5/0.5/3$ \\
Peak LR & $4{\times}10^{-4}$ & $2{\times}10^{-4}$ & $6{\times}10^{-5}$ \\
Tokens & ${\sim}5$B & ${\sim}10$B & ${\sim}60$B \\
\bottomrule
\end{tabular}}
\vspace{-0.2cm}
\end{table}

\paragraph{Models and datasets.}
{\method} uses a Llama-style pre-norm decoder~\citep{xiong2020layer,touvron2023llama} with grouped-query attention~\citep{ainslie2023gqa}, RoPE~\citep{su2024roformer} ($\theta=10^{4}$), and sparse SwiGLU MoE layers~\citep{shazeer2020glu,dai2024deepseekmoe,jiang2024mixtral}. \textbf{Each loop applies layers $\ell=1,2,\dots,M$ in order and repeats this walk for $H$ passes.} Models are pretrained on FineWeb-Edu~\citep{penedo2024fineweb} with sequence length $1{,}024$ and a $65{,}536$-token vocabulary. Table~\ref{tab:train} lists the architecture and optimization settings. The ${\sim}100$M model uses $M{=}6$, hidden size $384$, $24$ routed experts, and top-$k{=}6$. The ${\sim}350$M and ${\sim}1.7$B models use $30$ routed experts and $2$ shared experts, with top-$k{=}8$ comprising $6$ routed and $2$ shared experts. All scales use loop-specific sigmoid routers in float32 and share the same $z$-loss coefficient, capacity factor, and load-balancing bias learning rate. Loop-specific coefficients are fixed across scales ($\lambda{=}0.5$, $\beta{=}0.5$, $K{=}3$); width and expert dimensions scale with model size, while these coefficients do not. Within each scale, the settings in Table~\ref{tab:train} stay fixed when varying $H$, so the loop-depth comparisons in Section~\ref{sec:depth} isolate recurrence depth.

\paragraph{Hyperparameters.}
We train all models with AdamW~\citep{loshchilov2017decoupled} ($\beta_1=0.9$, $\beta_2=0.95$, weight decay $0.1$), gradient clipping at $1.0$, and a $10/80/10$ warmup--stable--decay schedule~\citep{hu2024minicpm} with a $0.1\times$ peak learning-rate floor. The global batch is $1{,}024$. 

\paragraph{Baselines.}
We compare {\method} with a non-looped $1{\times}$ baseline and three recurrent baselines matched in backbone, training data, and loop count. \ding{182}~``No tech'' loops the unmodified backbone without additional techniques. \ding{183}~``Res. scale'' builds on this native loop by applying only residual scaling. \ding{184}~``Embed inject'' builds on this native loop by applying only embedding re-injection. 

\paragraph{Zero-shot commonsense reasoning.}
We evaluate pretrained checkpoints with standard zero-shot prompts on seven tasks: PIQA~\citep{bisk2020piqa}, SIQA~\citep{sap2019social}, HellaSwag~\citep{zellers2019hellaswag}, WinoGrande~\citep{sakaguchi2021winogrande}, ARC-c and ARC-e~\citep{clark2018think}, and OBQA~\citep{mihaylov2018suitarmorconductelectricity}. Unless noted otherwise, we report length-normalized accuracy (\texttt{acc\_norm}) for OBQA, ARC-c, ARC-e, HellaSwag, and PIQA, and raw accuracy (\texttt{acc}) for WinoGrande and SIQA.

\subsection{Loop-depth experiments}
\label{sec:depth}

We evaluate loop depth under two protocols: near-iso-FLOP comparisons with a non-looped baseline (Section~\ref{sec:ablation}) and non-iso-FLOP sweeps with a fixed MoE layout, where effective depth and compute grow with $H$ (Section~\ref{sec:main}).

\begin{table}[!t]
\centering
\caption{Near-iso-FLOP comparison on the $M{=}10$, $E{=}80$ backbone. All models have 700M parameters and are trained on 10B tokens. Per-token layer cost is $f=H(6+3k)$ in units of $d^2$, anchored at $f_1=84$. Best results are bold.}
\label{tab:isoflops}
\vspace{0.1pt}
\small
\setlength{\tabcolsep}{2.6pt}
\resizebox{0.9\linewidth}{!}{%
\begin{tabular}{ccccc|c|cccccccc}
\toprule
Method & Loops & Eff. depth & top-$k$ & $f(H,k)$ & Val.ppl & OBQA & Wino & ARCC & ARCE & HSwg & SIQA & PIQA & Avg. \\
\midrule
\rowcolor{loopone}
Baseline & $1{\times}$ & $10$ & $26$ & $84$ & $18.36$ & $28.40$ & $51.38$ & $22.95$ & $43.10$ & $29.43$ & $35.72$ & $60.88$ & $38.84$ \\
{\method} & $2{\times}$ & $20$ & $12$ & $84$ & $17.37$ & $30.60$ & $49.33$ & $24.32$ & $44.57$ & $29.98$ & $34.65$ & $59.58$ & $39.00$ \\
{\method} & $3{\times}$ & $30$ & $8$ & $90$ & $16.91$ & $30.20$ & $50.38$ & $\mathbf{25.00}$ & $44.02$ & $30.05$ & $35.18$ & $60.55$ & $39.34$ \\
{\method} & $4{\times}$ & $40$ & $5$ & $84$ & $16.58$ & $30.20$ & $50.28$ & $24.74$ & $44.11$ & $30.14$ & $\mathbf{36.01}$ & $\mathbf{61.26}$ & $\mathbf{39.53}$ \\
{\method} & $5{\times}$ & $50$ & $4$ & $90$ & $\mathbf{16.54}$ & $\mathbf{30.80}$ & $50.20$ & $24.91$ & $\mathbf{44.70}$ & $\mathbf{30.46}$ & $34.65$ & $60.99$ & $\mathbf{39.53}$ \\
{\method} & $6{\times}$ & $60$ & $3$ & $90$ & $16.57$ & $30.40$ & $\mathbf{52.17}$ & $24.66$ & $44.02$ & $29.91$ & $35.16$ & $60.17$ & $39.50$ \\
\bottomrule
\end{tabular}}
\vspace{-0.3cm}
\end{table}

\subsubsection{Iso-FLOP comparison}
\label{sec:ablation}

We design a near-iso-FLOP experiment to test whether a looped {\method} outperforms a non-looped one at matched compute. The backbone has $M{=}10$ layers, hidden size $d{=}512$, expert width $I{=}d$, and a pool of $E{=}80$ routed experts (no shared experts); we vary $H$ and shrink top-$k$, while equivalent depth $MH$ still rises from $10$ to $60$. \textbf{These runs use full backpropagation through all loops, without segmented backpropagation}; Consequently, the comparison incurs no additional LM-head evaluations or backward passes associated with segmentation.  We count multiply--adds in units of $d^{2}$ per token per stacked layer. With $I{=}d$, grouped-query $qkv$ costs $3d^{2}$, causal SDPA costs $2d^{2}$, and $o_{\mathrm{proj}}$ costs $1d^{2}$, so attention is $6d^{2}$. Each routed SwiGLU expert is three $d{\times}d$ maps and costs $3d^{2}$; $k$ experts therefore cost $3kd^{2}$. Over $H$ loops the unit is
\begin{equation}
  f(H,k) = H(6+3k) = 3H(2+k).
  \label{eq:flopunit}
\end{equation}
We choose top-$k$ so that $f(H,k)\approx f_{1}=84$ (the non-looped anchor). Table~\ref{tab:isoflops} reports the compute budget of each loop count together with the experimental results. Val.ppl falls from the non-looped Baseline through Loop4, so four loops already improve on Loop2, Loop3, and Baseline (no-loop). Loop5 and Loop6 then stay with Loop4 rather than slipping. This already extends well beyond the two-loop setting adopted by prior looped models~\citep{gao2026loop,wang2026smelt,lab2026nanbeige4}.

\begin{table}[!t]
\centering
\caption{Non-iso-FLOP loop-depth sweeps at ${\sim}100$M and ${\sim}350$M scales, including partial stabilization baselines. We report validation perplexity ($\downarrow$) and zero-shot accuracy (\%, $\uparrow$). Best results within each scale are bold; $\dagger$ denotes an unrecoverable loss spike.}
\label{tab:hsweep}
\label{tab:partial}
\vspace{0.1pt}
\small
\setlength{\tabcolsep}{2.6pt}
\resizebox{0.95\linewidth}{!}{%
\begin{tabular}{ccccc|c|cccccccc}
\toprule
Model size & Method & Loops & Eff. depth & top-$k$ & Val.ppl & OBQA & Wino & ARCC & ARCE & HSwg & SIQA & PIQA & Avg. \\
\midrule
\rowcolor{loopone}
& Baseline & $1{\times}$ & $6$ & $6$ & $26.26$ & $27.20$ & $48.46$ & $22.95$ & $40.07$ & $26.85$ & $34.54$ & $57.02$ & $36.73$ \\
& No tech. & $3{\times}$ & $18$ & $6$ & $27.61$ & $27.80$ & $48.30$ & $21.93$ & $38.89$ & $26.24$ & $34.03$ & $56.96$ & $36.31$ \\
& No tech. & $6{\times}$ & $36$ & $6$ & $1311^{\dagger}$ & $27.40$ & $51.54$ & $24.49$ & $27.53$ & $25.51$ & $34.90$ & $48.91$ & $34.33^{\dagger}$ \\
& No tech. & $9{\times}$ & $54$ & $6$ & $1312^{\dagger}$ & $26.80$ & $47.75$ & $23.63$ & $27.27$ & $25.61$ & $34.90$ & $48.97$ & $33.56^{\dagger}$ \\
& No tech. & $12{\times}$ & $72$ & $6$ & $1587^{\dagger}$ & $27.20$ & $49.33$ & $24.06$ & $28.07$ & $25.34$ & $34.18$ & $49.51$ & $33.96^{\dagger}$ \\
& Embed inject & $3{\times}$ & $18$ & $6$ & $27.48$ & $24.80$ & $52.01$ & $22.27$ & $37.75$ & $26.21$ & $33.73$ & $54.68$ & $35.92$ \\
& Embed inject & $6{\times}$ & $36$ & $6$ & $71.28^{\dagger}$ & $24.60$ & $49.49$ & $21.84$ & $32.03$ & $24.79$ & $34.24$ & $52.72$ & $34.24^{\dagger}$ \\
& Embed inject & $9{\times}$ & $54$ & $6$ & $141.5^{\dagger}$ & $26.40$ & $50.99$ & $21.25$ & $31.31$ & $25.58$ & $34.44$ & $50.92$ & $34.41^{\dagger}$ \\
& Embed inject & $12{\times}$ & $72$ & $6$ & $186.4^{\dagger}$ & $25.80$ & $48.86$ & $22.61$ & $29.84$ & $25.03$ & $33.67$ & $51.37$ & $33.88^{\dagger}$ \\
& Res. scale & $3{\times}$ & $18$ & $6$ & $28.99$ & $26.80$ & $50.12$ & $23.29$ & $37.46$ & $26.64$ & $34.29$ & $56.37$ & $36.42$ \\
& Res. scale & $6{\times}$ & $36$ & $6$ & $34.67^{\dagger}$ & $25.20$ & $48.62$ & $25.17$ & $25.88$ & $25.84$ & $34.14$ & $48.59$ & $33.35^{\dagger}$ \\
& Res. scale & $9{\times}$ & $54$ & $6$ & $33.78^{\dagger}$ & $27.80$ & $47.75$ & $\mathbf{26.11}$ & $26.85$ & $25.42$ & $32.09$ & $49.89$ & $33.70^{\dagger}$ \\
& Res. scale & $12{\times}$ & $72$ & $6$ & $35.64^{\dagger}$ & $26.40$ & $48.54$ & $25.26$ & $26.64$ & $25.88$ & $33.52$ & $49.80$ & $33.72^{\dagger}$ \\
\rowcolor{loomrow}
& {\method} & $3{\times}$ & $18$ & $6$ & $23.21$ & $28.00$ & $49.49$ & $24.15$ & $39.81$ & $27.11$ & $33.21$ & $57.73$ & $37.07$ \\
\rowcolor{loomrow}
& {\method} & $6{\times}$ & $36$ & $6$ & $19.81$ & $29.20$ & $\mathbf{54.06}$ & $23.29$ & $42.30$ & $28.26$ & $\mathbf{35.57}$ & $59.90$ & $38.94$ \\
\rowcolor{loomrow}
& {\method} & $9{\times}$ & $54$ & $6$ & $19.29$ & $\mathbf{31.00}$ & $50.91$ & $24.32$ & $43.35$ & $\mathbf{28.75}$ & $35.26$ & $\mathbf{60.17}$ & $\mathbf{39.11}$ \\
\rowcolor{loomrow}
\multirow{-17}{*}{${\sim}100$M} & {\method} & $12{\times}$ & $72$ & $6$ & $\mathbf{19.26}$ & $30.80$ & $50.49$ & $25.17$ & $\mathbf{44.23}$ & $28.66$ & $35.21$ & $59.19$ & $\mathbf{39.11}$ \\
\midrule
\rowcolor{loopone}
& Baseline & $1{\times}$ & $10$ & $8$ & $20.07$ & $28.40$ & $48.54$ & $23.04$ & $42.42$ & $28.43$ & $34.80$ & $60.07$ & $37.96$ \\
& No tech. & $3{\times}$ & $30$ & $8$ & $23.75$ & $27.60$ & $51.54$ & $22.10$ & $40.07$ & $26.95$ & $34.75$ & $56.96$ & $37.14$ \\
& No tech. & $6{\times}$ & $60$ & $8$ & $1312^{\dagger}$ & $26.80$ & $49.96$ & $23.89$ & $27.53$ & $25.62$ & $35.31$ & $49.24$ & $34.05^{\dagger}$ \\
& No tech. & $9{\times}$ & $90$ & $8$ & $1313^{\dagger}$ & $26.60$ & $48.07$ & $24.32$ & $27.53$ & $25.69$ & $34.95$ & $49.46$ & $33.80^{\dagger}$ \\
& No tech. & $12{\times}$ & $120$ & $8$ & $1528^{\dagger}$ & $27.40$ & $49.41$ & $23.46$ & $26.92$ & $25.47$ & $34.29$ & $48.75$ & $33.67^{\dagger}$ \\
& Embed inject & $3{\times}$ & $30$ & $8$ & $42.76$ & $26.60$ & $50.99$ & $20.99$ & $35.27$ & $26.30$ & $33.52$ & $54.57$ & $35.46$ \\
& Embed inject & $6{\times}$ & $60$ & $8$ & $803.6^{\dagger}$ & $29.00$ & $50.67$ & $24.83$ & $27.82$ & $25.42$ & $34.60$ & $49.62$ & $34.57^{\dagger}$ \\
& Embed inject & $9{\times}$ & $90$ & $8$ & $780.6^{\dagger}$ & $28.00$ & $47.20$ & $25.34$ & $28.66$ & $25.55$ & $34.65$ & $49.95$ & $34.19^{\dagger}$ \\
& Embed inject & $12{\times}$ & $120$ & $8$ & $864.2^{\dagger}$ & $27.60$ & $49.09$ & $23.98$ & $29.41$ & $25.21$ & $33.98$ & $50.28$ & $34.22^{\dagger}$ \\
& Res. scale & $3{\times}$ & $30$ & $8$ & $23.80$ & $28.40$ & $50.99$ & $22.44$ & $40.07$ & $27.26$ & $34.44$ & $57.56$ & $37.31$ \\
& Res. scale & $6{\times}$ & $60$ & $8$ & $23.40$ & $27.60$ & $51.07$ & $22.35$ & $39.27$ & $27.67$ & $34.60$ & $56.47$ & $37.00$ \\
& Res. scale & $9{\times}$ & $90$ & $8$ & $33.89^{\dagger}$ & $24.80$ & $47.59$ & $24.66$ & $26.09$ & $25.53$ & $32.55$ & $48.59$ & $32.83^{\dagger}$ \\
& Res. scale & $12{\times}$ & $120$ & $8$ & $42.15^{\dagger}$ & $26.20$ & $50.83$ & $22.18$ & $34.93$ & $25.76$ & $34.90$ & $54.90$ & $35.67^{\dagger}$ \\
\rowcolor{loomrow}
& {\method} & $3{\times}$ & $30$ & $8$ & $18.27$ & $31.20$ & $\mathbf{51.70}$ & $24.15$ & $44.19$ & $29.04$ & $35.11$ & $60.28$ & $39.38$ \\
\rowcolor{loomrow}
& {\method} & $6{\times}$ & $60$ & $8$ & $15.35$ & $32.20$ & $50.43$ & $26.11$ & $47.98$ & $31.59$ & $36.13$ & $61.70$ & $40.88$ \\
\rowcolor{loomrow}
& {\method} & $9{\times}$ & $90$ & $8$ & $\mathbf{14.80}$ & $32.00$ & $50.12$ & $\mathbf{27.22}$ & $47.18$ & $32.41$ & $\mathbf{36.90}$ & $\mathbf{62.24}$ & $41.15$ \\
\rowcolor{loomrow}
\multirow{-17}{*}{${\sim}350$M} & {\method} & $12{\times}$ & $120$ & $8$ & $14.86$ & $\mathbf{32.40}$ & $51.14$ & $26.43$ & $\mathbf{48.23}$ & $\mathbf{32.78}$ & $36.80$ & $62.13$ & $\mathbf{41.42}$ \\
\bottomrule
\end{tabular}}
\vspace{-0.3cm}
\end{table}

\vspace{-0.3cm}
\subsubsection{Non-iso-FLOP loop experiments}
\label{sec:main}
\vspace{-0.1cm}

Table~\ref{tab:hsweep} supports two conclusions. \ding{182} \textbf{The complete recipe enables stable depth scaling.} Across both model scales, {\method} remains stable under deeper recurrence and improves validation perplexity and downstream accuracy over the width-matched non-looped baselines. \ding{183} \textbf{Relaxing the compute constraint favors deeper recurrence.} Although performance gains gradually saturate, the highest seven-task average is attained at 12 loops in this sweep, extending beyond the preferred depth under the near-iso-FLOP setting in Section~\ref{sec:ablation}.

Comparisons with the three recurrent baselines further clarify the importance of the complete recipe. \ding{182} \textbf{Native looping is insufficient.} ``No tech.'' loops the unmodified backbone without additional techniques and rapidly becomes unstable as recurrence deepens. \ding{183} \textbf{Residual scaling alone does not ensure stability.} ``Res. scale'' adds only residual scaling; it mitigates perplexity degradation but still encounters unrecoverable loss spikes at larger loop counts. \ding{184} \textbf{Embedding re-injection alone cannot sustain deep recurrence.} ``Embed inject'' adds only embedding re-injection and exhibits substantial degradation as loop depth increases. All three underperform the non-looped baselines even at shallow recurrence, supporting the effectiveness of the combined mechanisms in {\method} for translating additional recurrent computation into performance gains.

\vspace{-0.1cm}
\subsubsection{Scaling Up Training}
\label{sec:scaleup}
\vspace{-0.1cm}

To evaluate {\method} at a larger scale, we train a ${\sim}1.7$B-parameter model on ${\sim}60$B tokens. Table~\ref{tab:anneal} highlights two findings. \ding{182} Stability in large-scale training. The ${\sim}1.7$B-parameter {\method} remains stable over ${\sim}60$B training tokens. Nine loops perform best, at an unrolled scale of roughly $15$B parameters ($1.7\mathrm{B}\times 9$), achieving $7.77$ validation perplexity and $47.7\%$ average zero-shot accuracy. \ding{183} Scalability with model size. The same framework extends from ${\sim}100$M and ${\sim}350$M to ${\sim}1.7$B parameters without changing its loop-specific hyperparameters, while retaining the benefits of recurrent computation. At the ${\sim}1.7$B scale, nine loops reduce perplexity from $9.62$ to $7.77$ and improve average accuracy from $42.4\%$ to $47.7\%$ over the non-looped baseline. These results demonstrate the scalability of {\method} as MoE parameter counts increase.

\vspace{-0.3cm}
\begin{table}[!t]
\centering
\caption{Scaling {\method} to ${\sim}1.7$B parameters and ${\sim}60$B tokens. We report validation perplexity ($\downarrow$) and zero-shot accuracy (\%, $\uparrow$).}
\label{tab:anneal}
\vspace{0.1pt}
\small
\setlength{\tabcolsep}{3.2pt}
\resizebox{0.8\linewidth}{!}{%
\begin{tabular}{cccc|c|cccccccc}
\toprule
Method & Loops & Eff. depth & top-$k$ & Val.ppl & OBQA & Wino & ARCC & ARCE & HSwg & SIQA & PIQA & Avg. \\
\midrule
\rowcolor{loopone}
Baseline & $1{\times}$ & $15$ & $8$ & $9.62$ & $29.6$ & $50.3$ & $26.7$ & $50.2$ & $37.1$ & $37.2$ & $66.0$ & $42.4$ \\
{\method} & $3{\times}$ & $45$ & $8$ & $8.94$ & $31.6$ & $51.3$ & $27.2$ & $51.0$ & $39.1$ & $39.1$ & $68.0$ & $43.9$ \\
{\method} & $6{\times}$ & $90$ & $8$ & $7.91$ & $32.2$ & $51.5$ & $31.1$ & $57.0$ & $46.6$ & $39.5$ & $69.2$ & $46.7$ \\
{\method} & $9{\times}$ & $135$ & $8$ & $\mathbf{7.77}$ & $33.0$ & $\mathbf{53.6}$ & $\mathbf{31.7}$ & $\mathbf{57.7}$ & $\mathbf{48.8}$ & $39.2$ & $70.2$ & $\mathbf{47.7}$ \\
{\method} & $12{\times}$ & $180$ & $8$ & $7.84$ & $\mathbf{34.8}$ & $50.7$ & $30.5$ & $56.9$ & $48.6$ & $\mathbf{40.0}$ & $\mathbf{70.7}$ & $47.5$ \\
\bottomrule
\end{tabular}}
\vspace{-0.2cm}
\end{table}

\begin{table}[!t]
\centering
\caption{Component ablation of the ${\sim}350$M {\method} model with 9 loops, evaluated at training step 5,000. Each ablated variant removes one component from the full recipe; ``w/o routing refresh'' uses a shared router across loops, and ``w/o LR'' removes the Looping Residual. The best result in each column is shown in bold.}
\label{tab:ablate}
\vspace{0.1pt}
\small
\setlength{\tabcolsep}{3.2pt}
\resizebox{0.8\linewidth}{!}{%
\begin{tabular}{l|c|cccccccc}
\toprule
Variant & Val.ppl & OBQA & Wino & ARCC & ARCE & HSwg & SIQA & PIQA & Avg. \\
\midrule
\rowcolor{loomrow}
{\method} & $\mathbf{18.62}$ & $\mathbf{31.4}$ & $51.1$ & $\mathbf{26.1}$ & $43.3$ & $\mathbf{29.5}$ & $35.8$ & $61.4$ & $\mathbf{39.8}$ \\
w/o Res. scale & $24.63$ & $29.2$ & $50.1$ & $23.8$ & $42.6$ & $29.4$ & $35.9$ & $59.4$ & $38.6$ \\
w/o Embed inject & $26.05$ & $29.4$ & $50.2$ & $24.3$ & $43.9$ & $29.2$ & $36.2$ & $\mathbf{61.5}$ & $39.2$ \\
w/o MoE RMSNorm & $26.50$ & $29.6$ & $\mathbf{52.1}$ & $24.1$ & $43.6$ & $28.7$ & $35.5$ & $59.4$ & $39.0$ \\
w/o Looping Residual & $19.25$ & $30.8$ & $49.2$ & $24.2$ & $\mathbf{44.8}$ & $\mathbf{29.5}$ & $\mathbf{36.6}$ & $60.2$ & $39.3$ \\
w/o routing refresh & $19.83$ & $30.0$ & $51.7$ & $25.4$ & $43.1$ & $29.4$ & $35.7$ & $59.2$ & $39.2$ \\
\bottomrule
\end{tabular}}
\vspace{-0.3cm}
\end{table}

\vspace{-0.1cm}
\subsection{Ablation}
\label{sec:ablate350}
\vspace{-0.1cm}

\subsubsection{Component ablation}
\label{sec:ablate-comp}
\vspace{-0.1cm}

Table~\ref{tab:ablate} ablates components of the ${\sim}350$M, 9-loop {\method} model at step 5,000, supporting two design principles. \ding{182} Stable recurrent states are essential for deep looping. Removing residual scaling, embedding re-injection, or MoE RMSNorm substantially increases perplexity and reduces average downstream accuracy, underscoring the need to control residual accumulation, preserve input information, and regulate activation magnitudes under repeated block reuse. \ding{183} Effective recurrence benefits from computational diversity and information preservation. Removing per-loop routing or the Looping Residual worsens both metrics, supporting their roles in diversifying expert computation and retaining information across iterations. Together, these results support our central claim: scaling recurrent depth requires stable hidden states and useful computation from successive loops.

\subsubsection{Segmented backpropagation}
\label{app:bpk}

{\method} adopts segmented backpropagation with segment length $K{=}3$, detaching the hidden state between segments to restrict gradient propagation to the current segment. Table~\ref{tab:bpk} shows that segmentation improves training stability, validation perplexity, and average downstream accuracy relative to full backpropagation. The comparison holds the architecture, training data, optimizer, and global batch size fixed. For the deepest full-backpropagation runs, the training micro-batch is reduced to fit the memory budget; these runs are therefore not matched to the segmented runs in activation memory. By bounding the backward graph within each segment, segmentation limits activation storage while preserving forward information across loops, supporting stable training at greater recurrence depths. These findings motivate its use as the default training strategy in the non-iso-FLOP experiments.

\begin{table}[H]
\centering
\caption{Comparison of segmented backpropagation ($K{=}3$) and full backpropagation across loop depths on the ${\sim}350$M model.}
\label{tab:bpk}
\small
\setlength{\tabcolsep}{3pt}
\resizebox{0.7\linewidth}{!}{%
\begin{tabular}{ccc|cccccccc}
\toprule
Loops & Backprop & Val.ppl & OBQA & Wino & ARCC & ARCE & HSwg & SIQA & PIQA & Avg. \\
\midrule
& $K{=}3$ & $\mathbf{18.27}$ & $\mathbf{31.2}$ & $\mathbf{51.7}$ & $\mathbf{24.2}$ & $\mathbf{44.2}$ & $\mathbf{29.0}$ & $35.1$ & $60.3$ & $\mathbf{39.4}$ \\
\multirow{-2}{*}{$3{\times}$} & none & $26.63$ & $29.2$ & $50.2$ & $23.1$ & $43.7$ & $28.4$ & $\mathbf{35.2}$ & $\mathbf{60.8}$ & $38.7$ \\
\midrule
& $K{=}3$ & $\mathbf{15.35}$ & $\mathbf{32.2}$ & $50.4$ & $\mathbf{26.1}$ & $\mathbf{48.0}$ & $\mathbf{31.6}$ & $\mathbf{36.1}$ & $\mathbf{61.7}$ & $\mathbf{40.9}$ \\
\multirow{-2}{*}{$6{\times}$} & none & $25.33$ & $29.2$ & $\mathbf{51.5}$ & $24.7$ & $43.2$ & $29.0$ & $35.7$ & $60.3$ & $39.1$ \\
\midrule
& $K{=}3$ & $\mathbf{14.80}$ & $\mathbf{32.0}$ & $\mathbf{50.1}$ & $\mathbf{27.2}$ & $\mathbf{47.2}$ & $\mathbf{32.4}$ & $\mathbf{36.9}$ & $\mathbf{62.2}$ & $\mathbf{41.1}$ \\
\multirow{-2}{*}{$9{\times}$} & none & $81.04^{\dagger}$ & $25.2$ & $47.5$ & $20.6$ & $32.9$ & $25.1$ & $35.1$ & $53.4$ & $34.3^{\dagger}$ \\
\midrule
& $K{=}3$ & $\mathbf{14.86}$ & $\mathbf{32.4}$ & $51.1$ & $\mathbf{26.4}$ & $\mathbf{48.2}$ & $\mathbf{32.8}$ & $\mathbf{36.8}$ & $\mathbf{62.1}$ & $\mathbf{41.4}$ \\
\multirow{-2}{*}{$12{\times}$} & none & $27.52$ & $30.8$ & $\mathbf{51.8}$ & $23.0$ & $43.2$ & $28.3$ & $35.9$ & $59.8$ & $39.0$ \\
\bottomrule
\end{tabular}
\vspace{-0.4cm}
}
\end{table}

\vspace{-0.4cm}
\section{Conclusion}
\label{sec:conc}
\vspace{-0.2cm}

This paper presents {\method}, a scalable recipe for extending recurrent depth in MoE LLMs beyond the commonly used two-loop regime. Our analysis identifies hidden-state variance growth and expert selection collapse as key obstacles to deeper recurrence. {\method} addresses these obstacles through residual scaling and embedding re-injection to stabilize recurrent states, together with per-loop routers and a Looping Residual to diversify expert computation and preserve information across iterations. Experiments spanning approximately $100$M to $1.7$B parameters and up to $60$B training tokens show stable scaling to $9$--$12$ loops, with the $1.7$B model scaling to $9$ loops. Under matched training FLOPs, {\method} scales to five loops while consistently achieving lower loss, providing evidence that recurrent depth offers a compute-efficient scaling axis for MoE language models.

\bibliographystyle{report}
\bibliography{mybib}

@article{wang2025chain,
  title={Chain-of-experts: unlocking the communication power of mixture-of-experts models},
  author={Wang, Zihan and Pan, Rui and Yao, Jiarui and Csordas, Robert and Li, Linjie and Yin, Lu and Wu, Jiajun and Zhang, Tong and Li, Manling and Liu, Shiwei},
  journal={arXiv preprint arXiv:2506.18945},
  year={2025}
}

@article{csordas2024moeut,
  title={Moeut: Mixture-of-experts universal transformers},
  author={Csord{\'a}s, R{\'o}bert and Irie, Kazuki and Schmidhuber, J{\"u}rgen and Potts, Christopher and Manning, Christopher D},
  journal={Advances in Neural Information Processing Systems},
  volume={37},
  pages={28589--28614},
  year={2024}
}

@article{chen2026loopmoe,
  title={LoopMoE: Unifying Iterative Computation with Mixture-of-Experts for Language Modeling},
  author={Chen, Wenkai and Li, Tianshu and Huang, Wenyong and Yin, Yichun and Shang, Lifeng and Qin, Chengwei},
  journal={arXiv preprint arXiv:2606.04438},
  year={2026}
}

@article{lab2026nanbeige4,
  title={Nanbeige4. 2-3B: Unlocking Agentic Capabilities in a Compact Model},
  author={Lab, Nanbeige and Yang, Chen and Huang, Chengrui and Lan, Fufeng and Chen, Hanhui and Zhou, Hao and Song, Huatong and Cao, Jiaqi and Zhu, Jiaying and Niu, Jinlin and others},
  journal={arXiv preprint arXiv:2607.22083},
  year={2026}
}

@article{li2024mix,
  title={Mix-ln: Unleashing the power of deeper layers by combining pre-ln and post-ln},
  author={Li, Pengxiang and Yin, Lu and Liu, Shiwei},
  journal={arXiv preprint arXiv:2412.13795},
  year={2024}
}

@article{xie2025mhc,
  title={mhc: Manifold-constrained hyper-connections},
  author={Xie, Zhenda and Wei, Yixuan and Cao, Huanqi and Zhao, Chenggang and Deng, Chengqi and Li, Jiashi and Dai, Damai and Gao, Huazuo and Chang, Jiang and Yu, Kuai and others},
  journal={arXiv preprint arXiv:2512.24880},
  year={2025}
}

@article{lee2026sparse,
  title={Sparse layers are critical to scaling looped language models},
  author={Lee, Ryan and Biloki, Jacob and Hu, Edward J and May, Jonathan},
  journal={arXiv preprint arXiv:2605.09165},
  year={2026}
}

@article{gao2026loop,
  title={Loop the Loopies!},
  author={Gao, Zitian and Chen, Yilong and Xiao, Yihao and Yang, Xinyu and Tao, Ran and Zhou, Joey and Dai, Bryan},
  journal={arXiv preprint arXiv:2607.16051},
  year={2026}
}

@article{fu2026simply,
  title={Simply Stabilizing the Loop via Fully Looped Transformer},
  author={Fu, Rao and Yang, Zixuan and Zhang, Jiankun and Ma, Jing and Chen, Hechang and Li, Yu and Chang, Yi},
  journal={arXiv preprint arXiv:2605.18797},
  year={2026}
}

@article{wang2026residual,
  title={On the Residual Scaling of Looped Transformers: Stability and Transferability},
  author={Wang, Shaowen and Li, Bingrui and Zhang, Ge and Huang, Wenhao and Yan, Shen and Li, Jian},
  journal={arXiv preprint arXiv:2606.18524},
  year={2026}
}

@article{wang2026smelt,
  title={SMELT: Scaling Laws for Compute-Matched MoE Looped Transformers},
  author={Wang, Shaowen and Zhang, Ge and Luo, Kairong and Wu, Yuhao and Liu, Shaofan and Liu, Jiaheng and Huang, Wenhao and Yan, Shen and Li, Jian},
  journal={arXiv preprint arXiv:2609.01343},
  year={2026}
}

@article{movahedi2026fixed,
  title={Fixed-Point Reasoners: Stable and Adaptive Deep Looped Transformers},
  author={Movahedi, Sajad and Milovanovi{\'c}, Vera and Feigin, Shlomo Libo and Theus, Alexander and Hofmann, Thomas and Boeva, Valentina and Rusch, T Konstantin and Orvieto, Antonio},
  journal={arXiv preprint arXiv:2606.18206},
  year={2026}
}

@article{geiping2026scaling,
  title={Scaling up test-time compute with latent reasoning: A recurrent depth approach},
  author={Geiping, Jonas and McLeish, Sean and Jain, Neel and Kirchenbauer, John and Singh, Siddharth and Bartoldson, Brian and Kailkhura, Bhavya and Bhatele, Abhinav and Goldstein, Tom},
  journal={Advances in Neural Information Processing Systems},
  volume={38},
  pages={41340--41391},
  year={2025}
}

@article{dehghani2018universal,
  title={Universal transformers},
  author={Dehghani, Mostafa and Gouws, Stephan and Vinyals, Oriol and Uszkoreit, Jakob and Kaiser, {\L}ukasz},
  journal={arXiv preprint arXiv:1807.03819},
  year={2018}
}

@article{lan2019albert,
  title={Albert: A lite bert for self-supervised learning of language representations},
  author={Lan, Zhenzhong and Chen, Mingda and Goodman, Sebastian and Gimpel, Kevin and Sharma, Piyush and Soricut, Radu},
  journal={arXiv preprint arXiv:1909.11942},
  year={2019}
}

@inproceedings{saunshi2025reasoning,
  title={Reasoning with latent thoughts: On the power of looped transformers},
  author={Saunshi, Nikunj and Dikkala, Nishanth and Li, Zhiyuan and Kumar, Sanjiv and J Reddi, Sashank},
  booktitle={International Conference on Learning Representations},
  volume={2025},
  pages={14855--14881},
  year={2025}
}

@inproceedings{giannou2023looped,
  title={Looped transformers as programmable computers},
  author={Giannou, Angeliki and Rajput, Shashank and Sohn, Jy-yong and Lee, Kangwook and Lee, Jason D and Papailiopoulos, Dimitris},
  booktitle={International Conference on Machine Learning},
  pages={11398--11442},
  year={2023},
  organization={PMLR}
}

@inproceedings{fan2025looped,
  title={Looped transformers for length generalization},
  author={Fan, Ying and Du, Yilun and Ramchandran, Kannan and Lee, Kangwook},
  booktitle={International Conference on Learning Representations},
  volume={2025},
  pages={14502--14520},
  year={2025}
}

@article{jolicoeur2025less,
  title={Less is more: Recursive reasoning with tiny networks},
  author={Jolicoeur-Martineau, Alexia},
  journal={arXiv preprint arXiv:2510.04871},
  year={2025}
}

@misc{wang2025hierarchicalreasoningmodel,
      title={Hierarchical Reasoning Model}, 
      author={Guan Wang and Jin Li and Yuhao Sun and Xing Chen and Changling Liu and Yue Wu and Meng Lu and Sen Song and Yasin Abbasi Yadkori},
      year={2025},
      eprint={2506.21734},
      archivePrefix={arXiv},
      primaryClass={cs.AI},
      url={https://arxiv.org/abs/2506.21734}, 
}

@article{sun2026curse,
  title={The curse of depth in large language models},
  author={Sun, Wenfang and Song, Xinyuan and Li, Pengxiang and Yin, Lu and Zheng, Yefeng and Liu, Shiwei},
  journal={Advances in Neural Information Processing Systems},
  volume={38},
  pages={163104--163136},
  year={2025}
}

@inproceedings{xiong2020layer,
  title={On layer normalization in the transformer architecture},
  author={Xiong, Ruibin and Yang, Yunchang and He, Di and Zheng, Kai and Zheng, Shuxin and Xing, Chen and Zhang, Huishuai and Lan, Yanyan and Wang, Liwei and Liu, Tieyan},
  booktitle={International conference on machine learning},
  pages={10524--10533},
  year={2020},
  organization={PMLR}
}

@article{noci2022signal,
  title={Signal propagation in transformers: Theoretical perspectives and the role of rank collapse},
  author={Noci, Lorenzo and Anagnostidis, Sotiris and Biggio, Luca and Orvieto, Antonio and Singh, Sidak Pal and Lucchi, Aurelien},
  journal={Advances in Neural Information Processing Systems},
  volume={35},
  pages={27198--27211},
  year={2022}
}

@article{team2026attention,
  title={Attention residuals},
  author={Team, Kimi and Chen, Guangyu and Zhang, Yu and Su, Jianlin and Xu, Weixin and Pan, Siyuan and Wang, Yaoyu and Wang, Yucheng and Chen, Guanduo and Yin, Bohong and others},
  journal={arXiv preprint arXiv:2603.15031},
  year={2026}
}

@article{shazeer2017outrageously,
  title={Outrageously large neural networks: The sparsely-gated mixture-of-experts layer},
  author={Shazeer, Noam and Mirhoseini, Azalia and Maziarz, Krzysztof and Davis, Andy and Le, Quoc and Hinton, Geoffrey and Dean, Jeff},
  journal={arXiv preprint arXiv:1701.06538},
  year={2017}
}

@article{fedus2022switch,
  title={Switch transformers: Scaling to trillion parameter models with simple and efficient sparsity},
  author={Fedus, William and Zoph, Barret and Shazeer, Noam},
  journal={Journal of Machine Learning Research},
  volume={23},
  number={120},
  pages={1--39},
  year={2022}
}

@article{jiang2024mixtral,
  title={Mixtral of experts},
  author={Jiang, Albert Q and Sablayrolles, Alexandre and Roux, Antoine and Mensch, Arthur and Savary, Blanche and Bamford, Chris and Chaplot, Devendra Singh and Casas, Diego de las and Hanna, Emma Bou and Bressand, Florian and others},
  journal={arXiv preprint arXiv:2401.04088},
  year={2024}
}

@inproceedings{dai2024deepseekmoe,
  title={Deepseekmoe: Towards ultimate expert specialization in mixture-of-experts language models},
  author={Dai, Damai and Deng, Chengqi and Zhao, Chenggang and Xu, RX and Gao, Huazuo and Chen, Deli and Li, Jiashi and Zeng, Wangding and Yu, Xingkai and Wu, Yu and others},
  booktitle={Proceedings of the 62nd annual meeting of the association for computational linguistics (volume 1: Long papers)},
  pages={1280--1297},
  year={2024}
}

@article{touvron2023llama,
  title={Llama: Open and efficient foundation language models},
  author={Touvron, Hugo and Lavril, Thibaut and Izacard, Gautier and Martinet, Xavier and Lachaux, Marie-Anne and Lacroix, Timoth{\'e}e and Rozi{\`e}re, Baptiste and Goyal, Naman and Hambro, Eric and Azhar, Faisal and others},
  journal={arXiv preprint arXiv:2302.13971},
  year={2023}
}

@inproceedings{ainslie2023gqa,
  title={Gqa: Training generalized multi-query transformer models from multi-head checkpoints},
  author={Ainslie, Joshua and Lee-Thorp, James and De Jong, Michiel and Zemlyanskiy, Yury and Lebr{\'o}n, Federico and Sanghai, Sumit},
  booktitle={Proceedings of the 2023 conference on empirical methods in natural language processing},
  pages={4895--4901},
  year={2023}
}

@article{su2024roformer,
  title={Roformer: Enhanced transformer with rotary position embedding},
  author={Su, Jianlin and Ahmed, Murtadha and Lu, Yu and Pan, Shengfeng and Bo, Wen and Liu, Yunfeng},
  journal={Neurocomputing},
  volume={568},
  pages={127063},
  year={2024},
  publisher={Elsevier}
}

@article{shazeer2020glu,
  title={Glu variants improve transformer},
  author={Shazeer, Noam},
  journal={arXiv preprint arXiv:2002.05202},
  year={2020}
}

@article{penedo2024fineweb,
  title={The fineweb datasets: Decanting the web for the finest text data at scale},
  author={Penedo, Guilherme and Kydl{\'\i}{\v{c}}ek, Hynek and Lozhkov, Anton and Mitchell, Margaret and Raffel, Colin and Von Werra, Leandro and Wolf, Thomas and others},
  journal={Advances in Neural Information Processing Systems},
  volume={37},
  pages={30811--30849},
  year={2024}
}

@article{hu2024minicpm,
  title={Minicpm: Unveiling the potential of small language models with scalable training strategies},
  author={Hu, Shengding and Tu, Yuge and Han, Xu and He, Chaoqun and Cui, Ganqu and Long, Xiang and Zheng, Zhi and Fang, Yewei and Huang, Yuxiang and Zhao, Weilin and others},
  journal={arXiv preprint arXiv:2404.06395},
  year={2024}
}

@article{loshchilov2017decoupled,
  title={Decoupled weight decay regularization},
  author={Loshchilov, Ilya and Hutter, Frank},
  journal={arXiv preprint arXiv:1711.05101},
  year={2017}
}

@article{kaplan2020scaling,
  title={Scaling laws for neural language models},
  author={Kaplan, Jared and McCandlish, Sam and Henighan, Tom and Brown, Tom B and Chess, Benjamin and Child, Rewon and Gray, Scott and Radford, Alec and Wu, Jeffrey and Amodei, Dario},
  journal={arXiv preprint arXiv:2001.08361},
  year={2020}
}

@article{hoffmann2022training,
  title={Training compute-optimal large language models},
  author={Hoffmann, Jordan and Borgeaud, Sebastian and Mensch, Arthur and Buchatskaya, Elena and Cai, Trevor and Rutherford, Eliza and Casas, Diego de Las and Hendricks, Lisa Anne and Welbl, Johannes and Clark, Aidan and others},
  journal={arXiv preprint arXiv:2203.15556},
  year={2022}
}

@article{hao2024training,
  title={Training large language models to reason in a continuous latent space},
  author={Hao, Shibo and Sukhbaatar, Sainbayar and Su, DiJia and Li, Xian and Hu, Zhiting and Weston, Jason and Tian, Yuandong},
  journal={arXiv preprint arXiv:2412.06769},
  year={2024}
}

@misc{bisk2020piqa,
      title={PIQA: Reasoning about Physical Commonsense in Natural Language}, 
      author={Yonatan Bisk and Rowan Zellers and Ronan Le Bras and Jianfeng Gao and Yejin Choi},
      year={2019},
      eprint={1911.11641},
      archivePrefix={arXiv},
      primaryClass={cs.CL},
      url={https://arxiv.org/abs/1911.11641}, 
}

@inproceedings{sap2019social,
  title={Social IQa: Commonsense reasoning about social interactions},
  author={Sap, Maarten and Rashkin, Hannah and Chen, Derek and Le Bras, Ronan and Choi, Yejin},
  booktitle={Proceedings of the 2019 conference on empirical methods in natural language processing and the 9th international joint conference on natural language processing (EMNLP-IJCNLP)},
  pages={4463--4473},
  year={2019}
}

@inproceedings{zellers2019hellaswag,
  title={Hellaswag: Can a machine really finish your sentence?},
  author={Zellers, Rowan and Holtzman, Ari and Bisk, Yonatan and Farhadi, Ali and Choi, Yejin},
  booktitle={Proceedings of the 57th annual meeting of the association for computational linguistics},
  pages={4791--4800},
  year={2019}
}

@article{sakaguchi2021winogrande,
  title={Winogrande: An adversarial winograd schema challenge at scale},
  author={Sakaguchi, Keisuke and Bras, Ronan Le and Bhagavatula, Chandra and Choi, Yejin},
  journal={Communications of the ACM},
  volume={64},
  number={9},
  pages={99--106},
  year={2021},
  publisher={ACM New York, NY, USA}
}

@article{clark2018think,
  title={Think you have solved question answering? try arc, the ai2 reasoning challenge},
  author={Clark, Peter and Cowhey, Isaac and Etzioni, Oren and Khot, Tushar and Sabharwal, Ashish and Schoenick, Carissa and Tafjord, Oyvind},
  journal={arXiv preprint arXiv:1803.05457},
  year={2018}
}

@inproceedings{mihaylov2018suitarmorconductelectricity,
  title={Can a suit of armor conduct electricity? a new dataset for open book question answering},
  author={Mihaylov, Todor and Clark, Peter and Khot, Tushar and Sabharwal, Ashish},
  booktitle={Proceedings of the 2018 conference on empirical methods in natural language processing},
  pages={2381--2391},
  year={2018}
}

\end{document}